\documentclass[conference]{IEEEtran}
\IEEEoverridecommandlockouts
\usepackage{cite}
\usepackage{amsmath,amssymb,amsfonts}
\usepackage{algorithmic}
\usepackage{graphicx}
\usepackage{textcomp}
\usepackage{xcolor}
 \usepackage[misc]{ifsym}
\usepackage{subcaption}
\usepackage{booktabs}
\usepackage{multirow}
\usepackage{bm}
\usepackage[linesnumbered,ruled,vlined]{algorithm2e}
\usepackage[capitalize]{cleveref}
\usepackage{array} 
\usepackage{bbding}
\def\BibTeX{{\rm B\kern-.05em{\sc i\kern-.025em b}\kern-.08em
    T\kern-.1667em\lower.7ex\hbox{E}\kern-.125emX}}
\begin{document}

\title{Efficient Dynamic-NeRF Based Volumetric Video Coding with Rate Distortion Optimization}

\author{\IEEEauthorblockN{1\textsuperscript{st} Zhiyu Zhang}
\IEEEauthorblockA{\textit{Institute of Image Communication and Network Engineering} \\
\textit{Shanghai Jiao Tong University}\\
Shanghai, China \\
zhiyu-zhang@sjtu.edu.cn}
\and
\IEEEauthorblockN{2\textsuperscript{nd} Guo Lu \textsuperscript{\Letter}}
\IEEEauthorblockA{\textit{Institute of Image Communication and Network Engineering} \\
\textit{Shanghai Jiao Tong University}\\
Shanghai, China \\
luguo2014@sjtu.edu.cn}
\and
\IEEEauthorblockN{3\textsuperscript{rd} Huanxiong Liang}
\IEEEauthorblockA{\textit{College of Electronics and Information Engineering} \\
\textit{Sichuan University}\\
Chengdu, China\\
2020141480088@stu.scu.edu.cn}
\and
\IEEEauthorblockN{4\textsuperscript{th} Anni Tang}
\IEEEauthorblockA{\textit{Institute of Image Communication and Network Engineering} \\
\textit{Shanghai Jiao Tong University}\\
Shanghai, China \\
memory97@sjtu.edu.cn}
\and
\IEEEauthorblockN{5\textsuperscript{th} Qiang Hu}
\IEEEauthorblockA{\textit{Institute of Image Communication and Network Engineering} \\
\textit{Shanghai Jiao Tong University}\\
Shanghai, China \\
qiang.hu@sjtu.edu.cn}
\and
\IEEEauthorblockN{6\textsuperscript{th} Li Song \textsuperscript{\Letter}}
\IEEEauthorblockA{\textit{Institute of Image Communication and Network Engineering} \\
\textit{Shanghai Jiao Tong University}\\
Shanghai, China \\
song\_li@sjtu.edu.cn}
}

\maketitle

\begin{abstract}
Volumetric videos, benefiting from immersive 3D realism and interactivity, hold vast potential for various applications, while the tremendous data volume poses significant challenges for compression. Recently, NeRF has demonstrated remarkable potential in volumetric video compression thanks to its simple representation and powerful 3D modeling capabilities, where a notable work is ReRF. 
However, ReRF separates the modeling from compression process, resulting in suboptimal compression efficiency.
In contrast, in this paper, we propose a volumetric video compression method based on dynamic NeRF in a more compact manner. Specifically, we decompose the NeRF representation into the coefficient fields and the basis fields, incrementally updating the basis fields in the temporal domain to achieve dynamic modeling. Additionally, we perform end-to-end joint optimization on the modeling and compression process to further improve the compression efficiency. Extensive experiments demonstrate that our method achieves higher compression efficiency compared to ReRF on various datasets.
\end{abstract}


\section{Introduction}
\label{sec:intro}

Volumetric video is an innovative technique for visual representation, allowing viewers to observe from any perspective~\cite{han2020vivo}. Thanks to its powerful 3D realism and interactivity, volumetric video holds vast potential for applications in metaverse, virtual reality, and various other domains. However, the acquisition of volumetric video typically requires multiple cameras capturing from different angles, resulting in data volumes that can be several times larger than traditional 2D videos~\cite{schreer2019capture}, which poses significant challenges in terms of storage and transmission. Therefore, efficient compression techniques for volumetric videos are crucial.

\begin{figure}[htbp]
  \centering
  \includegraphics[width=1.0\linewidth]{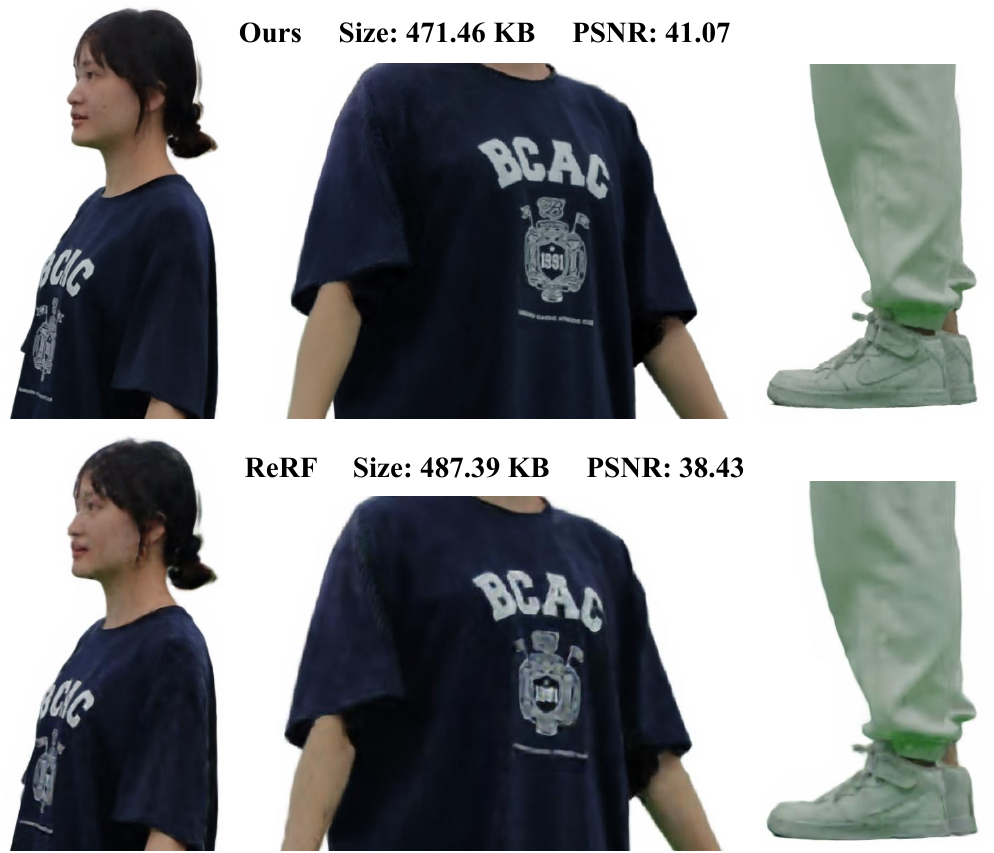}
  \vspace{-2mm}
  \caption{The rendering results of our method in comparison with ReRF.}
  \vspace{-5mm}
  \label{fig:demenstrate}
\end{figure}

Image-based methods~\cite{boyce2021mpeg} interpolate novel views within densely captured sequences and can be compressed by 2D video codecs. However, the quality of synthesized views in such methods is inferior to those based on 3D reconstruction. While geometry-based solutions~\cite{graziosi2020overview} involve the reconstruction and compression of dynamic point clouds, but they are vulnerable to occlusions and textureless regions. Recently, there has been a surge in employing Neural Radiance Fields (NeRF)~\cite{mildenhall2021nerf} to represent 3D scenes.
On one hand, NeRF utilizes a large-scale Multi-Layer Perceptron (MLP) to fit the colors and densities of points within the scene and employs volume rendering to generate images from arbitrary viewpoints. In other words, NeRF employs an MLP as an implicit representation of 3D scenes. On the other hand, compared to point clouds, NeRF can synthesize photo-realistic rendering results from novel viewpoints with richer details. 
To sum up, NeRF has the characteristics of compact representation and powerful 3D modeling capabilities, which is naturally suitable for volumetric video compression. 

Although MLP is a compact and powerful 3D representation, its high computational complexity poses significant challenges for applications that require real-time processing, such as volumetric video compression.
Consequently, some works~\cite{sun2022direct, muller2022instant, chen2022tensorf, chen2023dictionary} have focused on the utilization of explicit features like 3D grids~\cite{sun2022direct}, hash tables~\cite{muller2022instant} and tensors~\cite{chen2022tensorf} to accelerate the rendering of NeRF.
However, the utilization of explicit features also incurs additional storage consumption, which can be particularly catastrophic for dynamic 3D scenes. Many works~\cite{li2023compressing, 10402733, takikawa2022variable, shin2024binary, girish2023shacira, mahmoud2023cawa} have successfully achieved explicit feature compression of NeRF  without compromising rendering speed in static scenes. 
Recently, many works~\cite{fang2022fast, fridovich2023k, xian2021space, pumarola2021d, park2021nerfies, li2022neural} have extended NeRF to dynamic scenarios. However, there have been fewer efforts in compressing explicit features for dynamic scenes, with ReRF~\cite{wang2023neural} being a representative work in this area.

ReRF~\cite{wang2023neural} divides the dynamic neural radiance fields into equally-sized groups of feature grids (GOF). The first frame within each GOF is designated as an I-frame and is represented using a complete feature grid. Subsequent frames are referred to as P-frames, and ReRF employs a compact motion grid and residual grid for representation. Specifically, the motion grid of P-frames is used to warp the feature grid of the I-frame, while the residual grid compensates for information loss. Furthermore, ReRF introduces a set of compression methods tailored for the feature grid and residual grid, encoding them into smaller bitstreams.

However, accurately estimating the 3D motion grid proves to be a challenging task, and within ReRF~\cite{wang2023neural}, the motion grid is also downsampled to reduce storages. Consequently, the estimated motion grid in ReRF is coarse, making it hard to completely eliminate the redundancy of inter-frame radiance field features. Moreover, the modeling and compression process of the dynamic radiance field in ReRF are separate. In other words, ReRF first trains the radiance fields and then compresses them, lacking end-to-end optimization, which results in suboptimal compression efficiency.

In this paper, we propose a compact representation for dynamic neural radiance fields. Inspired by DiF~\cite{chen2023dictionary}, we decompose the radiance field representation into the coefficient fields and the basis fields. By incrementally updating the basis fields in the temporal domain, we achieve dynamic NeRF modeling. Additionally, we perform end-to-end optimization of both the modeling and compression process, thereby enhancing the rate-distortion performance. Since the quantization and entropy encoding operations in the compression process are non-differentiable, we introduce a method for simulating quantization and estimating bitrates, enabling gradient backpropagation. The introduction of simulated quantization and bitrate constraints during the radiance field modeling not only enhances the robustness of the trained radiance field features to quantization but also endows them with low-entropy. In summary, our contributions can be summarized as follows:
\begin{itemize}
\vspace{-1mm}
    \item We propose an efficient volumetric video compression method based on dynamic neural radiance fields, achieving state-of-the-art compression performance.
    \item We propose a modeling approach to compactly representing dynamic neural radiance fields, which effectively enhances compression efficiency.
    \item We devise a strategy that jointly optimizes the modeling and compression of dynamic NeRF, replacing non-differentiable quantization and entropy encoding with differentiable operations. This enables end-to-end joint optimization and enhances the rate-distortion(RD) performance.
\end{itemize}

\section{Method}

\begin{figure*}[htbp]
  \centering
  \includegraphics[width=0.9\linewidth]{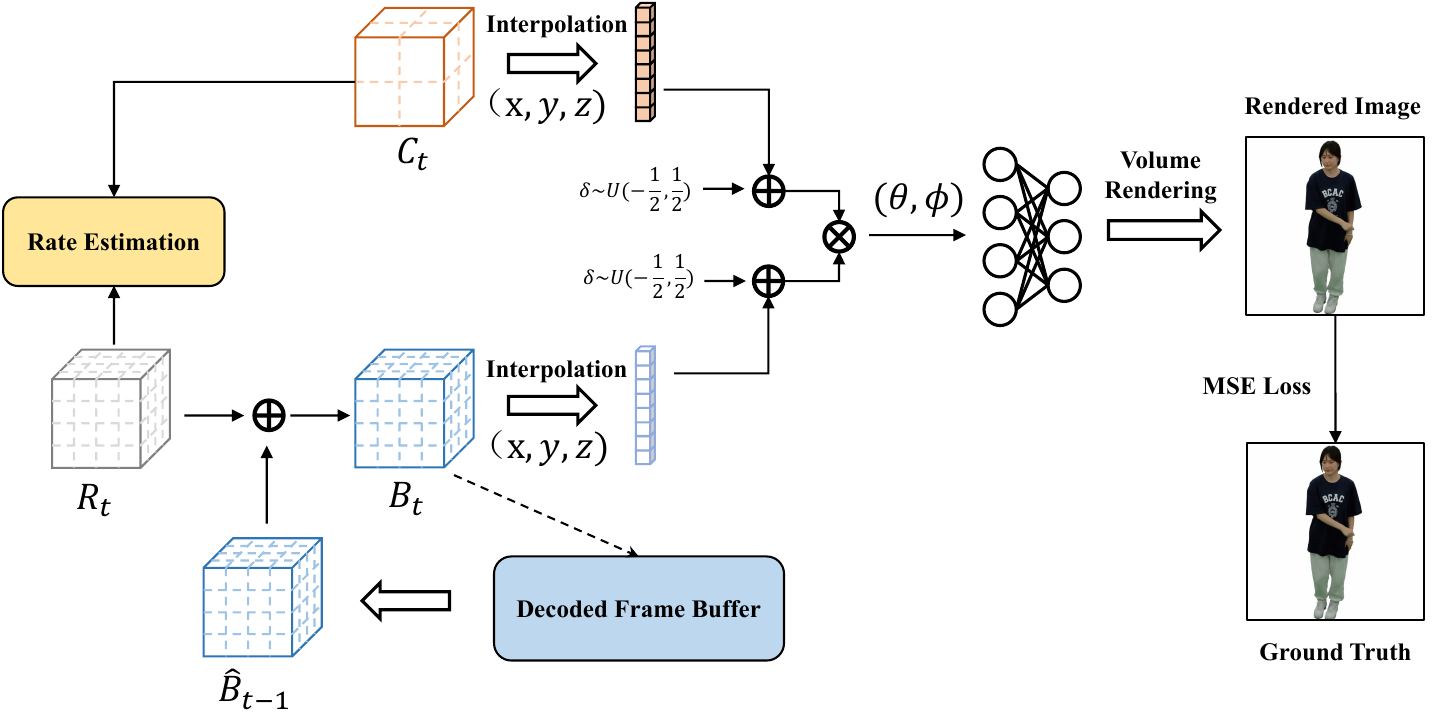}
  \caption{Training pipeline for the proposed method. (a) Initially, we load the basis field at the previous time step from the decoded frame buffer. During training, we only update the coefficient field and the residual field. (b) During end-to-end optimization, we estimate the rate of the coefficient field and the residual field as loss, using simulated quantization during the forward pass.}
  \vspace{-3mm}
  \label{fig:pipeline}
\end{figure*}

\subsection{Preliminaries}
\label{sec:representation}

NeRF~\cite{mildenhall2021nerf} learns a mapping function $g_\phi(\mathbf{x}, \mathbf{d}): \mathbb{R}^d \rightarrow \mathbb{R}^c$ that maps the coordinates $\mathbf{x}=(x, y, z)$ of points sampled along a ray $\mathbf{r}$, along with the viewing direction $\mathbf{d}=(\theta, \phi)$, to the corresponding color $\mathbf{c}$ and density $\sigma$:
\begin{equation}
\left(\mathbf{c}, \sigma \right) = g_\phi(\mathbf{x}, \mathbf{d})
\end{equation}
where the mapping function $g_\phi(\mathbf{x}, \mathbf{d})$ is fitted by a large-scale MLP. For volume rendering, the colors $\mathbf{c}_i$ and densities $\sigma_i$ of all sampled points along a ray $\mathbf{r}$ are accumulated to obtain the color $\hat{C}(\mathbf{r})$ of the corresponding pixel:
\begin{equation}
\begin{aligned}
\hat{C}(\mathbf{r}) &=\sum_{i=1}^N T_i \alpha_i \mathbf{c}_i &\\
T_i = \prod_{j=1}^{i-1}\left(1 - \alpha_i \right) & \quad
\alpha_i = 1 - \exp \left(- \sigma_i \delta_i \right) &
\end{aligned}
\end{equation}
where $T_i$ and $\alpha_i$ represent the transmittance and alpha value of the $i$-th sampled point and $\delta_i$ denotes the distance between adjacent sampled points.

To accelerate training and rendering, DiF~\cite{chen2023dictionary} decomposes the representation of NeRF into the coefficient fields and the basis fields. The basis fields capture the commonality of the signal, and the coefficient fields represent the spatial variation of the signal. Specifically, the coefficient fields are represented by a single-scale 3D grid $\mathbf{C}$, while the basis fields are represented by multi-scale 3D grids $\mathbf{B}$.

Initially, the coordinates of the sampled points $\mathbf{x}$ are utilized to obtain features through trilinear interpolation within the coefficient field $\mathbf{C}$ and basis field $\mathbf{B}$. Similar to the hash mapping technique employed in INGP~\cite{muller2022instant}, DiF incorporates a rawtooth coordinate transformation $\gamma$ when indexing the features of the basis field. Subsequently, the coefficient features $\mathbf{c}(\mathbf{x})$ and basis features $\mathbf{b}(\mathbf{x})$ are merged using the Hadamard product $\circ$, and finally mapped to color and density through a shallow MLP $\mathcal{P}$:

\begin{equation}
\begin{aligned}
\mathbf{c}(\mathbf{x}) &=\operatorname{interp}\left(\mathbf{x}, \mathbf{C} \right) \\
\mathbf{b}(\mathbf{x}) & =\operatorname{interp}\left(\gamma(\mathbf{x}), \mathbf{B}\right) \\
(\mathbf{c}, \sigma)  & = \mathcal{P}(\mathbf{c}(\mathbf{x}) \circ \mathbf{b}(\mathbf{x}), \mathbf{d})
\end{aligned}
\label{equ}
\end{equation}

\subsection{Compact Dynamic NeRF Representation}
\label{sec:dynamic}

When modeling dynamic neural radiance fields, in order to maintain efficient training and rendering, we employ the explicit representation similar to DiF~\cite{chen2023dictionary}. Specifically, for static radiance field, the representation consists of coefficient fields, basis fields, and a tiny MLP. Considering that the parameter of the MLP is negligible compared to the 3D grids, our main focus is on compressing the 3D grids.

\begin{table}[htbp]
\renewcommand\arraystretch{1.25}
\caption{{The bitrate allocation of different components in DiF. The basis grids occupy the majority bitrate of DiF, causing the necessity of fine compression.}}
\label{table:dif}
\centering
\begin{tabular}{cccc}
    \toprule[1.5pt]
     meta data  & MLP & coefficient grid &  basis grids \\
    \hline
    5.7\% & 0.4\%  & 10.3\% & 83.6\% \\
    \bottomrule[1.5pt]
\end{tabular}
\vspace{-3mm}
\end{table}

When extending the radiance field from static scenes to dynamic ones, the most naive approach would be to independently model the radiance field for each time step. In other words, at each time step $t$, both $\mathbf{C}_t$ and $\mathbf{B}_t$ are stored as individual representations.
However, such a method not only neglects the temporal continuity but also poses challenges for compression. Particularly for long-duration dynamic sequences, the storage requirements of individually modeling each time step are unacceptable. Therefore, we propose a compact modeling approach for dynamic radiance fields, which not only preserves temporal continuity but also facilitates compression.

Fig.~\ref{fig:pipeline} illustrates an overview of our proposed method. Initially, we decompose the representation of NeRF into coefficient fields and basis fields~\cite{chen2023dictionary}. When training the radiance fields for a new time step, we retrieve the previous frame's basis fields $\hat{\mathbf{B}}_{t-1}$ from the decoded frame buffer and maintain its unchanged throughout the current training process. During training, we only update the coefficient fields $\mathbf{C}_t$ and the residual fields $\mathbf{R}_t$ for the current frame. The coefficient fields represent the scene variation between the current time step and the previous one, while the residual fields compensate for the appearance of new regions in the current frame's scene. Once the training of the residual fields is complete, we add it to the previous frame's basis fields to obtain the basis fields for the current frame:
\begin{equation}
    \mathbf{B}_t =  \hat{\mathbf{B}}_{t-1} + \mathbf{R}_t
\end{equation}

Addtionally, to ensure temporal continuity and facilitate compression, we apply L1 regularization to the residual fields:
\begin{equation}
    \mathcal{L}_{reg} = \left\|\mathbf{R}_t\right\|_1
\end{equation}

With the aforementioned steps, we can represent the radiance field of a time step using $\mathbf{C}_t$ and $\mathbf{R}_t$. By adding $\mathbf{R}_t$ to $\hat{\mathbf{B}}_{t-1}$, we obtain $\mathbf{B}_t$. When rendering the image, we begin by sampling points in space and calculating the color and density of the sampled points using the equation described in Equ.~\ref{equ}. Subsequently, we perform volume rendering to generate images from arbitrary viewpoints.

\subsection{End-to-end Optimization}
\label{sec:end2end}

In this section, we introduce the end-to-end optimization method, which combines modeling and compression of dynamic radiance fields to further improve compression efficiency. Specifically, we incorporate simulated quantization and rate estimation in the modeling process, ensuring that the radiance field representation learned from modeling is robust to quantization and exhibits low entropy.

\vspace{0.2cm}
\noindent{\textbf{Simulated Quantization.}}
Quantization is a crucial operation in compression algorithms to reduce bit rates. However, it also introduces a certain degree of information loss. 
If we can incorporate the quantization operation in the process of radiance field modeling, the trained radiance field can effectively learn representations that are robust to quantization. 
Unfortunately, the gradient of the quantization function is almost zero everywhere, which means that incorporating quantization in the training process of radiance field modeling would hinder gradient descent optimization. Inspired by~\cite{balle_iclr}, we can introduce random uniform noise as a substitute for the quantization operation, simulating the information loss caused by quantization:
\begin{equation}
\hat{y} = \operatorname{round}(y) \quad \rightarrow \quad \tilde{y}=y+\delta
\end{equation}
where $\delta \sim \mathcal{U}\left[-\frac{1}{2}, \frac{1}{2}\right]$. 

During the compression process, we apply quantization to the 3D grid, thereby introducing information loss. In the modeling process, the 3D grid undergoes interpolation to generate features for subsequent forward processes. Thus, as depicted in Fig.~\ref{fig:pipeline}, we introduce uniform noise to the interpolated features $\mathbf{c}(x)$ and $\mathbf{b}(x)$ to simulate the quantization operation performed on the 3D grid. By simulating the quantization operation, we enhance the robustness of the radiacne field representation obtained through training, thereby mitigating the significant degradation in rendering quality that can arise from quantization.

\vspace{0.2cm}

\noindent{\textbf{Rate Estimation.}}
In compression algorithms, quantized parameters are further encoded into binary bitstreams using entropy encoding, and the size of bitstream represents the bitrate. 
Therefore, if we can obtain the true bitrate of the radiance field representation and use it as a loss function, we can optimize the radiance field representation for better compression performance. 
However, the process of entropy encoding is complex and non-differentiable, making it impossible to perform entropy encoding during the training process. From the theory of entropy encoding, we know that the theoretical compression upper bound of entropy encoding is the symbol's entropy. Therefore, we can estimate the entropy of the radiance field representation and use the estimated entropy as the loss function for optimization. This allows us to continuously reduce the entropy of the radiance field representation during the optimization process, endowing radiance field representation with low entropy.

Inspired by~\cite{mahmoud2023cawa, girish2023shacira}, we can approximate the probability mass function (PMF) of $\hat{y}$ by computing the cumulative distribution function (CDF) of $\tilde{y}$. Specifically, we assume that all parameters within the 3D grid follow a Laplace distribution. We establish two trainable parameters, $\mu$ and $b$, to represent the mean and scale of the Laplace distribution, respectively, continuously updating them during training. Consequently, the PMF of the quantized parameters can be approximated as follows:
\begin{equation}
    P(\hat{y}_i) = P_{cdf} \left( \tilde{y}_i + \frac{1}{2} \right) - P_{cdf} \left( \tilde{y}_i - \frac{1}{2} \right)
\end{equation}
where $\tilde{y}$ represents the simulated quantized radiance field representation during training, while $\hat{y}$ denotes the actual quantized radiance field representation. Then, we can estimate the rate loss by:
\begin{equation}
\mathcal{L}_{rate} = \frac{1}{N} \sum_{y_i \in \mathbf{F}} -\log_2\left[ P(\hat{y}_i)\right]
\end{equation}

In sunmmury, the total loss function can be written as:

\begin{equation}
\mathcal{L}_{total}=\sum_{\mathbf{r} \in \mathbb{R}}\|C(\mathbf{r})-\hat{C}(\mathbf{r})\|^2+\lambda_1 \mathcal{L}_{rate} + \lambda_2 \mathcal{L}_{reg}
\end{equation}

The first term represents the L2 loss between the image rendered by NeRF and the ground truth image, indicating the level of distortion. The parameter $\lambda_1$ is a rate-distortion trade-off parameter that controls the degree of compression. It allows for the balance between the rate loss and the distortion loss, determining the level of compression. The parameter $\lambda_2$ represents the degree of regularization applied to $\mathbf{R}_t$.

\section{Experiments}
\subsection{Implementation Details}

\noindent{\textbf{Datasets.}}
In this paper, we conducted experiments on the ReRF dataset~\cite{wang2023neural} and the Dna-rendering dataset~\cite{cheng2023dna}. The ReRF dataset consists of three sequences, comprising a total of 74 perspectives, with a resolution of $1920 \times 1080$. As for the Dna-rendering dataset, we selected four sequences, encompassing a total of 60 perspectives. Given the varying resolutions of the images in the Dna-rendering dataset across different perspectives, we performed a central crop and downsampling to achieve a uniform resolution of  $900 \times 600$ for each perspective. For each sequence, we set aside four perspectives as the test set, while the remaining perspectives were allocated for the training set.

\noindent{\textbf{Setups.}}
The parameters of the Laplace distribution were initialized with $\mu$ set to 0 and $b$ set to 0.01. 
For loss function, we configured four different $\lambda_1$: 0.0001, 0.0005, 0.002, and 0.005, where each value corresponds to a different compression ratio. Additionally, we set the weight $\lambda_2$ for the regularization loss to 0.0001. 
During the experiment, we inserted an I-frame every 20 frames, which was represented by coefficient fields $\mathbf{C}$ and basis fields $\mathbf{B}$.

\begin{figure}
    \centering
    
    \begin{subfigure}{0.23\textwidth}
        \centering
        \includegraphics[width=\linewidth]{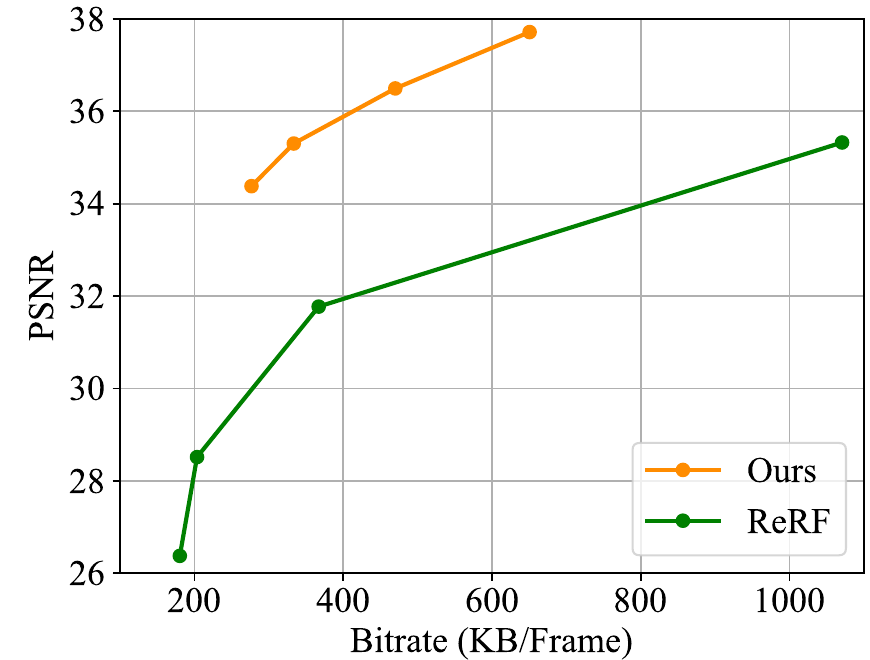}
        \caption{Dna-rendering train}
        \label{fig:image1}
    \end{subfigure}
    \hfill
    \begin{subfigure}{0.23\textwidth}
        \centering
        \includegraphics[width=\linewidth]{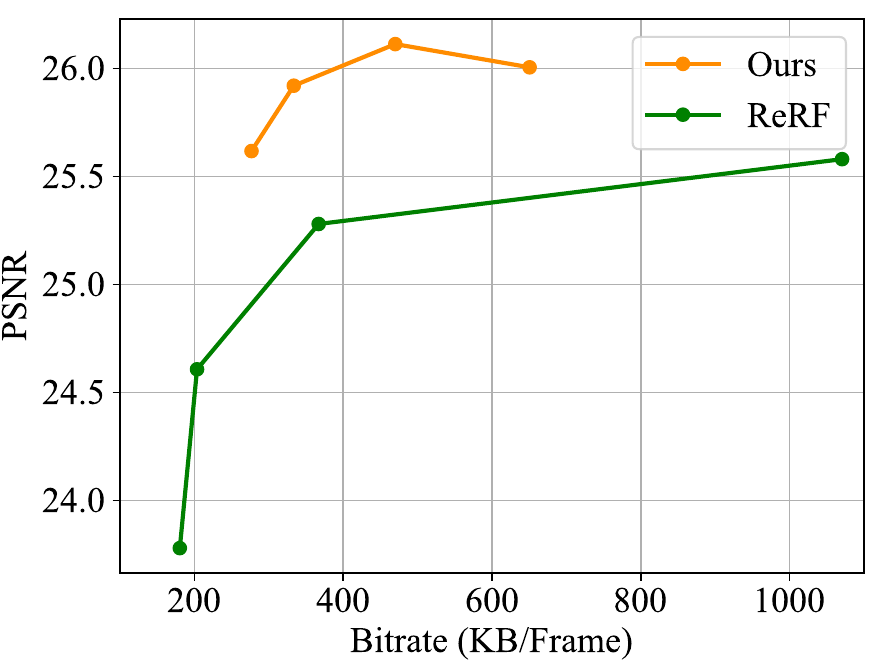}
        \caption{Dna-rendering test}
        \label{fig:image2}
    \end{subfigure}
    
    \vspace{0.5cm} 
    
    \begin{subfigure}{0.23\textwidth}
        \centering
        \includegraphics[width=\linewidth]{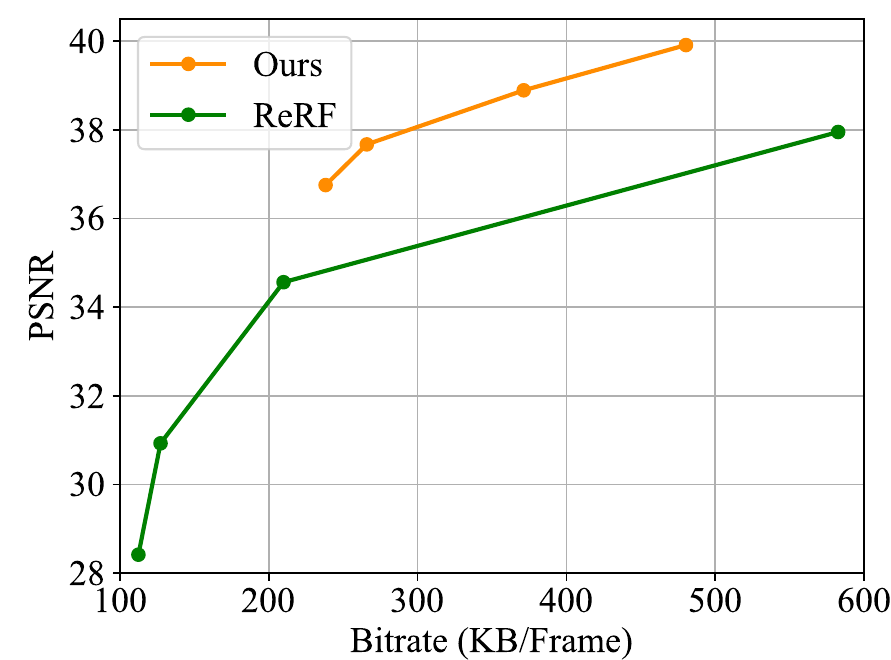}
        \caption{ReRF train}
        \label{fig:image3}
    \end{subfigure}
    \hfill
    \begin{subfigure}{0.23\textwidth}
        \centering
        \includegraphics[width=\linewidth]{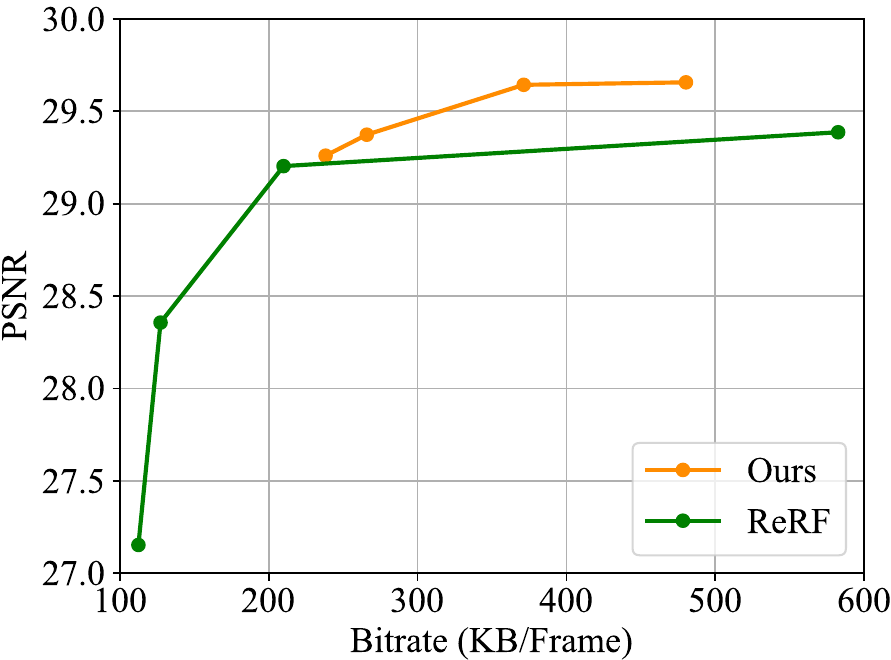}
        \caption{ReRF test}
        \label{fig:image4}
    \end{subfigure}
    
    \caption{The performance comparison of ReRF and our method in both \textit{ReRF} and \textit{Dna-rendering} dataset.
    \textbf{BD-rate}: \textit{Dna-rendering train}:\textbf{-66.54\%}, \textit{Dna-rendering test}:\textbf{-28.71\%}, \textit{ReRF train}:\textbf{-45.50\%}, \textit{ReRF test}:\textbf{-33.24\%}.}
    \label{fig:rd_curve}
\end{figure}


\begin{figure}[t]
  \centering
  \includegraphics[width=1.0\linewidth]{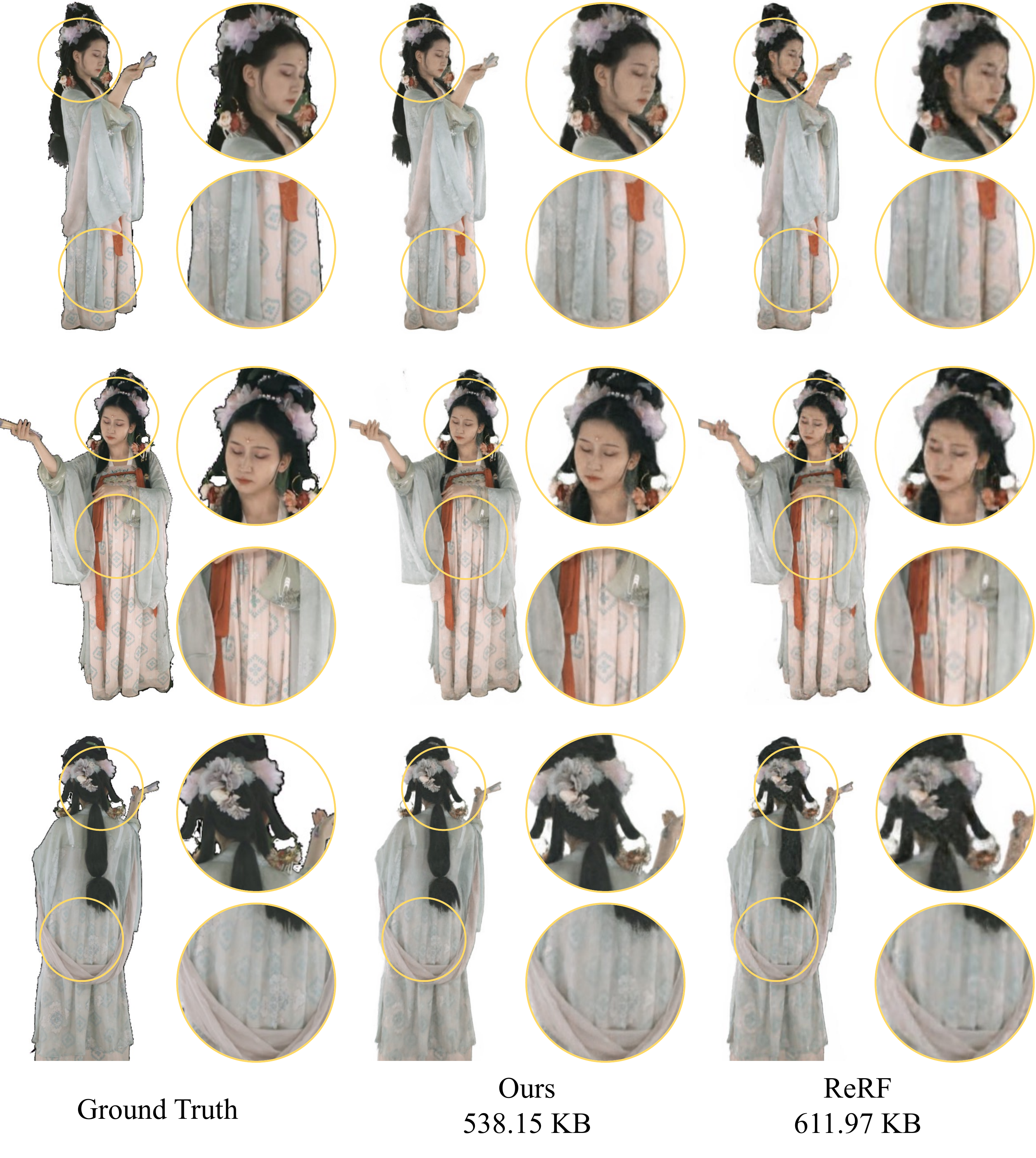}
  \vspace{-2mm}
  \caption{Comparing the multi-view rendering results of our approach and ReRF on the DNA-rendering sequence, under similar bitrate.}
  \vspace{-5mm}
  \label{fig:compare}
\end{figure}

\subsection{Experimental Results}

In the comparative experiments, we compared our method against the state-of-the-art approach for 3D video compression based on NeRF, known as ReRF~\cite{wang2023neural}. ReRF employs a similar method to JPEG for compressing model parameters, so we conducted tests on ReRF using four quality parameters.

For quantitative comparison, we utilized the PSNR metric to evaluate the rendering quality. Fig.~\ref{fig:rd_curve} illustrates the quantitative results on the Dna-rendering dataset and ReRF dataset. From Fig.~\ref{fig:rd_curve}, it is evident that our method outperforms ReRF in terms of rate-distortion performance on both datasets. On the training set of Dna-rendering, the BD-rate is $\textbf{-66.54\%}$, and on the testing set, the BD-rate is $\textbf{-28.71\%}$. For ReRF dataset, the BD-rate on the training set is $\textbf{-45.50\%}$, and on the test set, the BD-rate is $\textbf{-33.24\%}$. Our performance gain stems from two aspects: firstly, our proposed dynamic NeRF representation is inherently more compact, and secondly, by introducing simulated quantization and bitrate constraints during training, the NeRF representation possesses low entropy, making it more amenable to compression. 

\begin{table}[t]
\renewcommand\arraystretch{1.1}
\caption{Step-by-step Analysis. The 2nd row shows the results of adding dynamic modeling to the baseline, and the 3rd row shows the results of the full model.}
\label{table:ablation}
\centering
\begin{tabular}{c|ccc}
    \toprule[1.5pt]
       & Bitrate & PSNR &  PSNR  \\
       & (Per Frame) & (Train) & (Test) \\
    \hline
    baseline & 18.48 MB  & 42.75 & 30.59 \\
    + dynamic modeling  & 1222.13 KB  & 41.82 & 30.38 \\
    + joint optimization & 471.46 KB & 41.07 & 30.37 \\
    \bottomrule[1.5pt]
\end{tabular}
\vspace{-3mm}
\end{table}

Fig.~\ref{fig:compare} shows the visual results of a sequence from the Dna-rendering dataset. We also provide more visual results in the supplemental material. Compared to ReRF, our method achieves a more realistic and immersive rendering of the scene while consuming less storage. Specifically, our method excels in rendering clearer details, such as facial features and the accessories worn. As depicted in Fig.~\ref{fig:compare}, the images rendered by ReRF exhibit severe distortions in facial features and lack details in the accessories. In contrast, our method accurately renders the facial features and captures intricate details of the accessories, showcasing its superior performance.

\subsection{Ablation Studies}

We conducted ablation studies on the kpop sequence in the ReRF dataset~\cite{wang2023neural} to validate the effectiveness of our method.

\textbf{Baseline.}
We consider DiF~\cite{chen2023dictionary} as the baseline for our method. However, while DiF is designed for static scenes, when extending it to dynamic scenes, we employ DiF to model each time step individually. Additionally, to ensure a fair comparison, we utilize the 7zip which is a lossless compression algorithm to compress the DiF models.

In this paper, we propose two enhanced methods for volumetric video compression based on dynamic NeRF. The first method is a compact dynamic NeRF modeling approach, while the second method involves end-to-end optimization of dynamic NeRF modeling and compression. To validate the effectiveness of these two methods, we conducted step-by-step ablation studies, progressively incorporating these two techniques on top of the baseline. 

In the first ablation study, we employed our proposed dynamic modeling method on the baseline approach, performing dynamic modeling and subsequently quantizing and entropy encoding the obtained representations. In the second ablation study, building upon the first one, we introduced simulated quantization and bitrate constraints during the training phase to achieve end-to-end optimization.

The results of the ablation studies are listed in Table~\ref{table:ablation}. Compared to the baseline, our dynamic modeling method achieves a compression ratio of approximately 15 times while only incurring a minimal loss in PSNR. Furthermore, introducing end-to-end optimization further compresses the representation of dynamic NeRF without significantly impacting PSNR. The results of the ablation studies demonstrate the effectiveness of our proposed methods.

\section{Conclusion}
In this paper, we present an efficient dynamic-NeRF based volumetric video coding method.
We propose a compact modeling approach for representing dynamic NeRF. Specifically, we decompose the radiance field into the coefficient field and the basis field, and incrementally update the basis field in the temporal domain to achieve dynamic modeling. Furthermore, we employ an end-to-end optimization for both modeling and compression, introducing simulated quantization and bitrate constraints during the modeling process. Experimental results demonstrate that our approach outperforms the state-of-the-art method, ReRF, in terms of compression efficiency on two challenging datasets. 
The modeling and compression method we propose, based on dynamic neural radiance fields, allows for significant reduction in the data size of volumetric videos. This provides a fundamental basis for the widespread application of volumetric videos.

\section*{Acknowledgment}
This work was supported by National Key R\&D Project of China(2019YFB1802701), National Natural Science Foundation of China(62102024), the Fundamental Research Funds for the Central Universities; in part by the 111 project under Grant B07022 and Sheitc No.150633; in part by the Shanghai Key Laboratory of Digital Media Processing and Transmissions.

\bibliographystyle{IEEEtran}
\bibliography{reference}

\begin{thebibliography}{10}
\providecommand{\url}[1]{#1}
\csname url@samestyle\endcsname
\providecommand{\newblock}{\relax}
\providecommand{\bibinfo}[2]{#2}
\providecommand{\BIBentrySTDinterwordspacing}{\spaceskip=0pt\relax}
\providecommand{\BIBentryALTinterwordstretchfactor}{4}
\providecommand{\BIBentryALTinterwordspacing}{\spaceskip=\fontdimen2\font plus
\BIBentryALTinterwordstretchfactor\fontdimen3\font minus \fontdimen4\font\relax}
\providecommand{\BIBforeignlanguage}[2]{{%
\expandafter\ifx\csname l@#1\endcsname\relax
\typeout{** WARNING: IEEEtran.bst: No hyphenation pattern has been}%
\typeout{** loaded for the language `#1'. Using the pattern for}%
\typeout{** the default language instead.}%
\else
\language=\csname l@#1\endcsname
\fi
#2}}
\providecommand{\BIBdecl}{\relax}
\BIBdecl

\bibitem{han2020vivo}
B.~Han, Y.~Liu, and F.~Qian, ``Vivo: Visibility-aware mobile volumetric video streaming,'' in \emph{Proceedings of the 26th annual international conference on mobile computing and networking}, 2020, pp. 1--13.

\bibitem{schreer2019capture}
O.~Schreer, I.~Feldmann, S.~Renault, M.~Zepp, M.~Worchel, P.~Eisert, and P.~Kauff, ``Capture and 3d video processing of volumetric video,'' in \emph{2019 IEEE International conference on image processing (ICIP)}.\hskip 1em plus 0.5em minus 0.4em\relax IEEE, 2019, pp. 4310--4314.

\bibitem{boyce2021mpeg}
J.~M. Boyce, R.~Dor{\'e}, A.~Dziembowski, J.~Fleureau, J.~Jung, B.~Kroon, B.~Salahieh, V.~K.~M. Vadakital, and L.~Yu, ``Mpeg immersive video coding standard,'' \emph{Proceedings of the IEEE}, vol. 109, no.~9, pp. 1521--1536, 2021.

\bibitem{graziosi2020overview}
D.~Graziosi, O.~Nakagami, S.~Kuma, A.~Zaghetto, T.~Suzuki, and A.~Tabatabai, ``An overview of ongoing point cloud compression standardization activities: Video-based (v-pcc) and geometry-based (g-pcc),'' \emph{APSIPA Transactions on Signal and Information Processing}, vol.~9, p. e13, 2020.

\bibitem{mildenhall2021nerf}
B.~Mildenhall, P.~P. Srinivasan, M.~Tancik, J.~T. Barron, R.~Ramamoorthi, and R.~Ng, ``Nerf: Representing scenes as neural radiance fields for view synthesis,'' \emph{Communications of the ACM}, vol.~65, no.~1, pp. 99--106, 2021.

\bibitem{sun2022direct}
C.~Sun, M.~Sun, and H.-T. Chen, ``Direct voxel grid optimization: Super-fast convergence for radiance fields reconstruction,'' in \emph{Proceedings of the IEEE/CVF Conference on Computer Vision and Pattern Recognition}, 2022, pp. 5459--5469.

\bibitem{muller2022instant}
T.~M{\"u}ller, A.~Evans, C.~Schied, and A.~Keller, ``Instant neural graphics primitives with a multiresolution hash encoding,'' \emph{ACM Transactions on Graphics (ToG)}, vol.~41, no.~4, pp. 1--15, 2022.

\bibitem{chen2022tensorf}
A.~Chen, Z.~Xu, A.~Geiger, J.~Yu, and H.~Su, ``Tensorf: Tensorial radiance fields,'' in \emph{European Conference on Computer Vision}.\hskip 1em plus 0.5em minus 0.4em\relax Springer, 2022, pp. 333--350.

\bibitem{chen2023dictionary}
A.~Chen, Z.~Xu, X.~Wei, S.~Tang, H.~Su, and A.~Geiger, ``Dictionary fields: Learning a neural basis decomposition,'' \emph{ACM Transactions on Graphics (TOG)}, vol.~42, no.~4, pp. 1--12, 2023.

\bibitem{li2023compressing}
L.~Li, Z.~Shen, Z.~Wang, L.~Shen, and L.~Bo, ``Compressing volumetric radiance fields to 1 mb,'' in \emph{Proceedings of the IEEE/CVF Conference on Computer Vision and Pattern Recognition}, 2023, pp. 4222--4231.

\bibitem{10402733}
Z.~Zhang, A.~Tang, C.~Zhu, G.~Lu, R.~Xie, and L.~Song, ``High-fidelity free-view talking head synthesis for low-bandwidth video conference,'' in \emph{2023 IEEE International Conference on Visual Communications and Image Processing (VCIP)}, 2023, pp. 1--5.

\bibitem{takikawa2022variable}
T.~Takikawa, A.~Evans, J.~Tremblay, T.~M{\"u}ller, M.~McGuire, A.~Jacobson, and S.~Fidler, ``Variable bitrate neural fields,'' in \emph{ACM SIGGRAPH 2022 Conference Proceedings}, 2022, pp. 1--9.

\bibitem{shin2024binary}
S.~Shin and J.~Park, ``Binary radiance fields,'' \emph{Advances in Neural Information Processing Systems}, vol.~36, 2024.

\bibitem{girish2023shacira}
S.~Girish, A.~Shrivastava, and K.~Gupta, ``Shacira: Scalable hash-grid compression for implicit neural representations,'' in \emph{Proceedings of the IEEE/CVF International Conference on Computer Vision}, 2023, pp. 17\,513--17\,524.

\bibitem{mahmoud2023cawa}
O.~Mahmoud, T.~Ladune, and M.~Gendrin, ``Cawa-nerf: Instant learning of compression-aware nerf features,'' \emph{arXiv preprint arXiv:2310.14695}, 2023.

\bibitem{fang2022fast}
J.~Fang, T.~Yi, X.~Wang, L.~Xie, X.~Zhang, W.~Liu, M.~Nie{\ss}ner, and Q.~Tian, ``Fast dynamic radiance fields with time-aware neural voxels,'' in \emph{SIGGRAPH Asia 2022 Conference Papers}, 2022, pp. 1--9.

\bibitem{fridovich2023k}
S.~Fridovich-Keil, G.~Meanti, F.~R. Warburg, B.~Recht, and A.~Kanazawa, ``K-planes: Explicit radiance fields in space, time, and appearance,'' in \emph{Proceedings of the IEEE/CVF Conference on Computer Vision and Pattern Recognition}, 2023, pp. 12\,479--12\,488.

\bibitem{xian2021space}
W.~Xian, J.-B. Huang, J.~Kopf, and C.~Kim, ``Space-time neural irradiance fields for free-viewpoint video,'' in \emph{Proceedings of the IEEE/CVF Conference on Computer Vision and Pattern Recognition}, 2021, pp. 9421--9431.

\bibitem{pumarola2021d}
A.~Pumarola, E.~Corona, G.~Pons-Moll, and F.~Moreno-Noguer, ``D-nerf: Neural radiance fields for dynamic scenes,'' in \emph{Proceedings of the IEEE/CVF Conference on Computer Vision and Pattern Recognition}, 2021, pp. 10\,318--10\,327.

\bibitem{park2021nerfies}
K.~Park, U.~Sinha, J.~T. Barron, S.~Bouaziz, D.~B. Goldman, S.~M. Seitz, and R.~Martin-Brualla, ``Nerfies: Deformable neural radiance fields,'' in \emph{Proceedings of the IEEE/CVF International Conference on Computer Vision}, 2021, pp. 5865--5874.

\bibitem{li2022neural}
T.~Li, M.~Slavcheva, M.~Zollhoefer, S.~Green, C.~Lassner, C.~Kim, T.~Schmidt, S.~Lovegrove, M.~Goesele, R.~Newcombe \emph{et~al.}, ``Neural 3d video synthesis from multi-view video,'' in \emph{Proceedings of the IEEE/CVF Conference on Computer Vision and Pattern Recognition}, 2022, pp. 5521--5531.

\bibitem{wang2023neural}
L.~Wang, Q.~Hu, Q.~He, Z.~Wang, J.~Yu, T.~Tuytelaars, L.~Xu, and M.~Wu, ``Neural residual radiance fields for streamably free-viewpoint videos,'' in \emph{Proceedings of the IEEE/CVF Conference on Computer Vision and Pattern Recognition}, 2023, pp. 76--87.

\bibitem{balle_iclr}
\BIBentryALTinterwordspacing
J.~Ball\'{e}, V.~Laparra, and E.~P. Simoncelli, ``End-to-end optimized image compression,'' in \emph{Int'l Conf on Learning Representations (ICLR)}, Toulon, France, April 2017, available at http://arxiv.org/abs/1611.01704. [Online]. Available: \url{https://arxiv.org/abs/1611.01704}
\BIBentrySTDinterwordspacing

\bibitem{cheng2023dna}
W.~Cheng, R.~Chen, S.~Fan, W.~Yin, K.~Chen, Z.~Cai, J.~Wang, Y.~Gao, Z.~Yu, Z.~Lin \emph{et~al.}, ``Dna-rendering: A diverse neural actor repository for high-fidelity human-centric rendering,'' in \emph{Proceedings of the IEEE/CVF International Conference on Computer Vision}, 2023, pp. 19\,982--19\,993.

\end{thebibliography}

\end{document}